%% file: main.tex
\documentclass[10pt,twocolumn,letterpaper]{article}
\usepackage{etoolbox}
\newtoggle{arxiv}
\toggletrue{arxiv}
\iftoggle{arxiv}{
    \usepackage[pagenumbers]{wacv}
}{
    \usepackage{wacv}
}

\input{preamble}
\definecolor{wacvblue}{rgb}{0.21,0.49,0.74}
\usepackage[pagebackref,breaklinks,colorlinks,allcolors=wacvblue]{hyperref}

\newcommand{\customfootnotetext}[2]{%
  \begingroup
  \setlength{\skip\footins}{0pt}
  \renewcommand{\thefootnote}{#1}%
  \footnotetext{#2}%
  \endgroup
}

\title{\OURS{} \faFlag: Surface-Based Framework for Aerodynamic Simulation with 3D Gaussians}

\author{
Hongru Yan$^{1,}$\textsuperscript{*}
\quad Xiang Zhang$^1$
\quad Zeyuan Chen$^1$
\quad Fangyin Wei$^2$
\quad Zhuowen Tu$^1$\\
$^1$UC San Diego \quad $^2$Princeton University
}

\begin{document}
\twocolumn[{%
	\renewcommand\twocolumn[1][]{#1}%
	\maketitle
	\begin{center}
	\includegraphics[width=\linewidth,trim=2em 0 2em 0,clip]{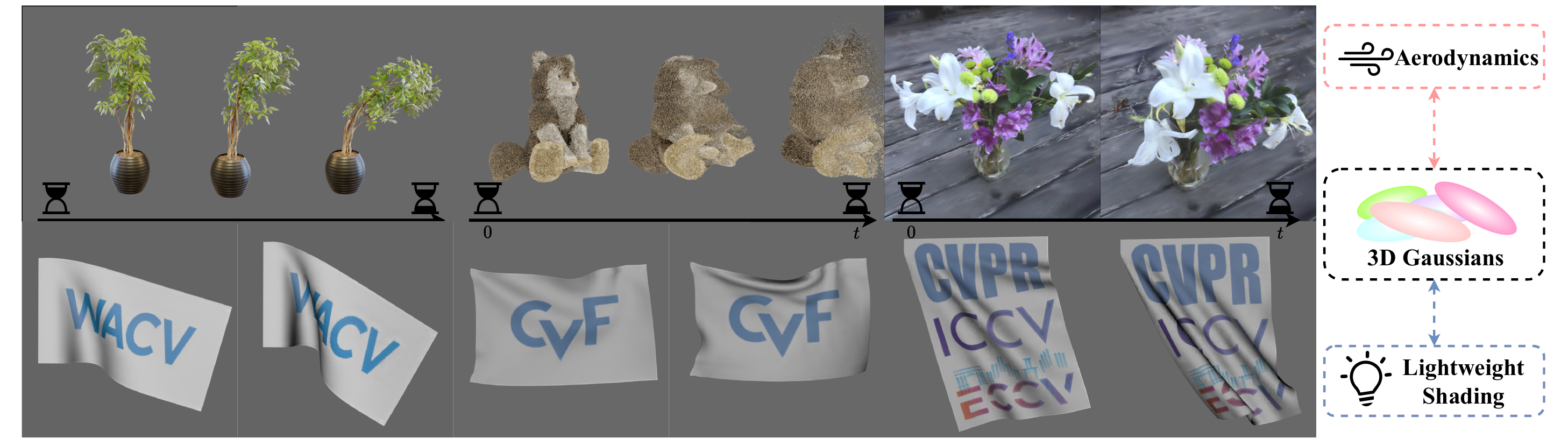}
	\captionof{figure}{
	\OURS{} is a unified surface-based framework that couples aerodynamic simulation with lightweight shading for efficient and realistic dynamics.
    }
    \vspace{1em}
    \label{fig:teaser}
    \vspace{1em}
	\end{center}
}]
\customfootnotetext{*}{Work done during internship at UC San Diego.}

\input{sec/0_abstract}
\input{sec/1_intro}
\input{sec/2_related}
\input{sec/3_method}
\input{sec/4_experiments}

\input{sec/5_conclusion}
\input{sec/acknowledgment}

{
    \small
	\bibliographystyle{ieeenat_fullname}
	\bibliography{main}
}

\iftoggle{arxiv}{%
    \clearpage
    \appendix

    \counterwithin{figure}{section}
    \counterwithin{table}{section}
    \counterwithin{equation}{section}
    \setcounter{table}{0}
    \setcounter{figure}{0}
    \setcounter{equation}{0}
    \section{Appendix}

    \let\origsection\section
    \let\origsubsection\subsection
    \let\origsubsubsection\subsubsection
    \let\section\subsection
    \let\subsection\subsubsection
  
    \input{sec/supplementary}
}{}

\end{document}

%% file: preamble.tex
\usepackage[table]{xcolor}
\usepackage{bm}
\usepackage[ruled,vlined]{algorithm2e}
\usepackage{tikz}
\usepackage{algorithmic}
\usepackage{fontawesome5}
\usepackage{multirow}
\usepackage{amssymb}
\usepackage{marvosym}
\usepackage{wrapfig}

\usetikzlibrary{fit,calc}
\newcolumntype{H}{>{\setbox0=\hbox\bgroup}c<{\egroup}@{}}
\newlength{\dhatheight}

\newcommand\OURS{Gaussian Swaying}

\definecolor{commentcolor}{RGB}{110,154,155}   

\SetKwInput{KwInput}{Input}
\SetKwInput{KwOutput}{Output}

%% file: sec/0_abstract.tex
\begin{abstract}

Branches swaying in the breeze, flags rippling in the wind, and boats rocking on the water all show how aerodynamics shape natural motion -- an effect crucial for realism in vision and graphics. In this paper, we present Gaussian Swaying, a surface-based framework for aerodynamic simulation using 3D Gaussians. Unlike mesh-based methods that require costly meshing, or particle-based approaches that rely on discrete positional data, Gaussian Swaying models surfaces continuously with 3D Gaussians, enabling efficient and fine-grained aerodynamic interaction. Our framework unifies simulation and rendering on the same representation: Gaussian patches, which support force computation for dynamics while simultaneously providing normals for lightweight shading. Comprehensive experiments on both synthetic and real-world datasets across multiple metrics demonstrate that Gaussian Swaying achieves state-of-the-art performance and efficiency, offering a scalable approach for realistic aerodynamic scene simulation.

\end{abstract}

%% file: sec/1_intro.tex
\section{Introduction}

Wind-driven motion is a defining aspect of our daily visual experience: branches sway, flags billow, and boats rock on the water. Capturing such dynamics is essential for realism in computer vision and graphics~\cite{runia2020cloth}. Traditional simulation techniques such as the Material Point Method (MPM)~\cite{sulsky1995application,hu2018moving} or Position-Based Dynamics (PBD)~\cite{macklin2016xpbd,muller2007position} are versatile across materials, but they incur high computational costs and struggle to meet the demand for fast yet visually faithful simulation. Moreover, because wind interactions are fundamentally surface-driven, these methods typically require explicit mesh reconstruction~\cite{jiang2017anisotropic,deul2018direct}, which complicates simulation for complex or evolving structures.

Recent advances in neural scene representations, such as NeRF~\cite{mildenhall2020nerf} and Gaussian Splatting~\cite{kerbl3Dgaussians}, have opened new possibilities for efficient dynamic simulation~\cite{xu2022deforming,yuan2022nerf,li2023pacnerf,feng2024pie,xie2024physgaussian,zhong2025reconstruction}. For instance, PhysGaussian~\cite{xie2024physgaussian} integrates Gaussian kernels with MPM particles to accelerate simulation and rendering. However, such approaches remain particle-based, modeling dynamics as collections of discrete points~\cite{rukhovich2021fcaf3d}. While efficient, they lack explicit surface awareness, limiting their ability to capture fine-grained aerodynamic interactions, where surface normals and areas directly govern how forces are applied.

In this work, we introduce \OURS{}, a surface-based framework for aerodynamic simulation with 3D Gaussians. Unlike particle-only or mesh-based approaches, \OURS{} treats Gaussian kernels as continuous surface patches characterized by normals and areas, enabling direct aerodynamic force computation. The same representation supports lightweight shading tied to surface properties, producing a realistic appearance that adapts naturally to lighting and view direction. By unifying simulation and rendering on a single surface-based representation, \OURS{} eliminates costly intermediate meshing steps, improving both efficiency and fidelity.

We validate \OURS{} through extensive experiments on synthetic and real-world datasets, including a custom flag simulation benchmark. Results show that \OURS{} achieves state-of-the-art performance and efficiency for aerodynamics, providing a practical and scalable solution for realistic aerodynamic simulation in vision and graphics.

Our key contributions are summarized as follows:

\begin{itemize}
    \item \textbf{Surface-based aerodynamic framework.} We propose \OURS{}, a novel framework for simulating aerodynamics with 3D Gaussians, eliminating costly meshing while supporting efficient, visually realistic simulation.

    \item \textbf{Surface-aware simulation and rendering.} We formulate Gaussians as continuous surface patches, rather than discrete particles, to unify aerodynamic force computation with lightweight shading for realism.

    \item \textbf{Extensive evaluation.} We conduct comprehensive experiments on both synthetic and real-world datasets, including a new flag simulation benchmark, demonstrating state-of-the-art accuracy and efficiency.
\end{itemize}

%% file: sec/2_related.tex
\section{Related Work}

\subsection{Neural Scene Representations}  

Neural scene representations have advanced rapidly for geometry modeling and downstream vision tasks. Some approaches focus on explicit 3D representations such as point clouds~\cite{yang2019pointflow,xu2024bayesian}, voxels~\cite{choy20163d}, and meshes~\cite{wang2018pixel2mesh,siddiqui2024meshgpt,zhang2025vertexregen}. In contrast, implicit representations define continuous functions over space: Signed Distance Fields (SDFs)~\cite{liu2020dist,zhang2023uni,zhao2025depr} model geometry via a continuous distance function, while Neural Radiance Fields (NeRF)~\cite{mildenhall2020nerf} extend implicit modeling to volumetric radiance, enabling high-quality novel-view synthesis. However, NeRF relies on fully connected networks and massive ray queries, resulting in slow training and inference. Subsequent works have improved its efficiency~\cite{fridovich2022plenoxels,muller2022instant} and extended its capabilities to high-fidelity synthesis~\cite{xu2022point}, multi-view generalization~\cite{liu2022neuray,chen2021mvsnerf,zheng2024oponerf}, and scene understanding~\cite{liu2023semantic,mirzaei2023spin,chen2024mvip}.  

More recently, 3D Gaussian Splatting~\cite{kerbl3Dgaussians} has emerged as an efficient alternative, directly representing scenes with explicit Gaussian primitives for real-time rendering. This representation has been applied to surface reconstruction~\cite{jiang2024gaussianshader,guedon2023sugar,chen:2024:vcrgaus}, object detection~\cite{yan2025gaussiandet}, semantic understanding~\cite{shi2023language}, and editing~\cite{chen2024gaussianeditor,tao2025c3editor}. Notably, works such as GaussianShader~\cite{jiang2024gaussianshader}, SuGaR~\cite{guedon2023sugar}, and Gaussian-Det~\cite{yan2025gaussiandet} highlight the strength of Gaussians in capturing surface-level details. These advances motivate our use of 3D Gaussians as surface representations, extending them beyond static reconstruction or recognition toward dynamic aerodynamic simulation.

\subsection{Dynamic Scenes and Simulations}  

Modeling dynamic scenes with neural representations such as NeRF and 3D Gaussians has become an active area in computer vision~\cite{lin2025omniphysgs,borycki2024gasp,shao2024gaussim}. Several works extend NeRF by incorporating time as an additional parameter to capture non-rigid motion~\cite{D-nerf,park2021nerfies}, while others achieve efficient dynamic view synthesis with Gaussian-based representations~\cite{duan20244d}. Beyond view synthesis, researchers have also explored simulation. NeRF-Editing~\cite{yuan2022nerf} and Deforming-NeRF~\cite{xu2022deforming} extract meshes or deformation cages from NeRFs, while PAC-NeRF~\cite{li2023pacnerf} integrates the material point method (MPM) with particles, and PIE-NeRF~\cite{feng2024pie} introduces mesh-less least squares kernels. PhysGaussian~\cite{xie2024physgaussian} represents a milestone by connecting Gaussian kernels with MPM particles for realistic and efficient physics simulation, followed by Spring-Gaus~\cite{zhong2025reconstruction}, which employs a spring–mass model for system identification.  

However, existing Gaussian-based simulation methods treat Gaussian kernels purely as positional primitives for physics engines, without explicitly modeling surfaces. This limits their ability to capture fine-grained aerodynamic interactions that are inherently surface-driven. Our work addresses this gap by formulating Gaussians as continuous surface patches and incorporating them directly into the simulation process, enabling smoother and more realistic surface dynamics.

\subsection{Aerodynamic Simulation} 

Aerodynamic simulation has traditionally relied on physics-based methods such as the finite element method (FEM), the material point method (MPM)~\cite{jiang2016material,hu2018moving,sulsky1995application}, and position-based dynamics (PBD)~\cite{macklin2016xpbd,muller2007position}. While these approaches are versatile across different physical settings, achieving real-time wind-driven dynamics with high visual fidelity remains a long-standing challenge in computer vision and graphics~\cite{han2019hybrid,stomakhin2013material,jiang2017anisotropic,lv2022efficient}. Moreover, such methods typically require extensive manual preparation, including mesh construction and vertex constraint tuning, which makes them less suitable for automatic or complex dynamic scenes. Incorporating realistic shading effects~\cite{cook1984distributed,barron2014shape} further increases computational cost, limiting their practicality for efficient simulation–rendering pipelines.

Motivated by these limitations, we propose a surface-based yet mesh-free framework that directly leverages 3D Gaussians as continuous surface representations. Our method eliminates costly meshing, reduces manual setup, and integrates lightweight shading, enabling efficient aerodynamic simulation with compelling visual realism.

%% file: sec/3_method.tex
\section{Method}

\begin{figure*}[!ht]
\centering
\includegraphics[width=\linewidth,trim=3em 0 2em 0,clip]{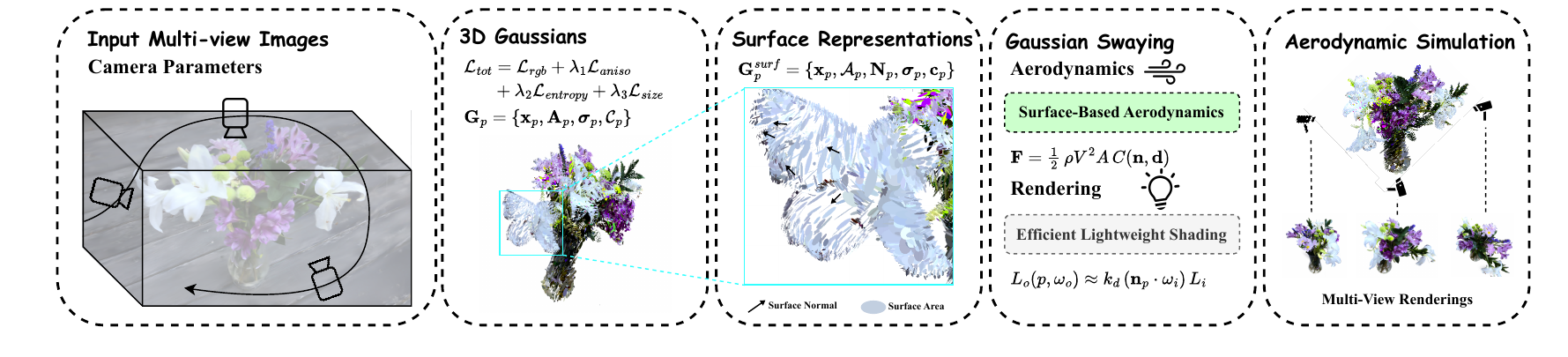}
\caption{
\textbf{Pipeline Overview.} \OURS{} is a unified surface-based framework for aerodynamics. By representing 3D Gaussians as continuous surface patches, it integrates aerodynamic simulation and realistic rendering within the same representation. Gaussian patches support aerodynamic force computation while also providing normals for lightweight shading, enabling realistic and efficient visual effects. Multi-view renderings of simulation results are shown on the right.
}
\label{fig:workflow}
\end{figure*}

We present \OURS{}, a unified surface-based framework for aerodynamics, with an overview illustrated in \cref{fig:workflow}. Our approach formulates 3D Gaussians as surface patches, enabling aerodynamic simulation and realistic rendering within the same representation. In this section, we first review the relevant preliminaries, then describe the proposed formulation in detail, including surface-aware Gaussian modeling, aerodynamic simulation, and lightweight shading.

\subsection{Preliminaries}

\paragraph{Continuum Mechanics.}  
Continuum mechanics describes the motion of a deformable body via a deformation map $\mathbf{x} = \phi(\mathbf{X},t)$ from material space $\Omega^0$ to world space $\Omega^t$ at time $t$. The deformation gradient $\mathbf{F} = \frac{\partial \phi}{\partial \mathbf{X}}(\mathbf{X},t)$ characterizes local rotation, stretch, and shear~\cite{bonet1997nonlinear}. $\mathbf{F}$ is updated via conservation of mass and momentum, which is given by:
\begin{equation}
    \frac{D\rho}{Dt} + \rho\nabla\cdot\mathbf{v} = 0
\end{equation}

\begin{equation}
    \rho\frac{D\mathbf{v}}{Dt} = \nabla\cdot\boldsymbol{\sigma} + \mathbf{f}^\mathrm{ext}
\end{equation}
where $\mathbf{f}^\mathrm{ext}$ is the external force and $\boldsymbol{\sigma}$ is the Cauchy stress.

\paragraph{Material Point Method.}
The material point method solves the above conservation laws by combining Eulerian grids with Lagrangian particles~\cite{jiang2016material,stomakhin2013material}. It is widely used for simulating diverse materials~\cite{yue2015continuum,ram2015material,klar2016drucker,jiang2015affine}. MPM alternates between two stages: (i) Particle-to-Grid (P2G), where momentum and mass are transferred from particles to the grid, and (ii) Grid-to-Particle (G2P), where updated velocities are mapped back:
\begin{equation}
    \frac{m_i}{\Delta t}(\mathbf{v}_i^{n+1} - \mathbf{v}_i^n) = - \sum\limits_p \tau_p^n \nabla w_{ip}^n V_p^0 + \mathbf{f}_i^\mathrm{ext}
\end{equation}
where $i$ and $p$ denote grid nodes and particles, respectively. $\tau_p = \det(\mathbf{F}_p)\boldsymbol{\sigma}_p$ is the Kirchhoff stress, $V_p^0$ is the initial particle volume, and $w_{ip}^n$ is the B-spline kernel evaluated at particle position $\mathbf{x}_p^n$. Further algorithmic details can be found in supplementary materials.

\paragraph{Gaussian Splatting.} 3D Gaussian Splatting~\cite{kerbl3Dgaussians} represents a scene using Gaussian primitives for real-time rendering. Each Gaussian $\mathbf{G}_p = \{\mathbf{x}_p, \mathbf{A}_p, \boldsymbol{\sigma}_p, \mathcal{C}_p\}$ is parameterized by position $\mathbf{x}_p$, covariance matrix $\mathbf{A}_p$, opacity $\boldsymbol{\sigma}_p$, and spherical harmonic (SH) coefficients $\mathcal{C}_p$. The Gaussian color $\mathbf{c}_p$ is derived from $\mathcal{C}_p$ and the view direction, and the final pixel color is obtained through $\alpha$-blending:
\begin{equation}
\label{eq:alpha_blending}
C = \sum\limits^n_{i=1}\mathbf{c}_i\alpha_i \prod_{j=1}^{i-1} (1 - \alpha_j)
\end{equation}
where $n$ is the number of Gaussians. Parameters of Gaussians are optimized via back-propagation during training.

\subsection{\OURS{}}
\subsubsection{Surface Formulation}

Unlike PhysGaussian~\cite{xie2024physgaussian} and other point-based methods that treat Gaussians as volumetric particles, \OURS{} adopts a surface-based representation. Point-based formulations are effective for general dynamics but lack explicit surface awareness, which is essential for aerodynamic interactions.

We define a Gaussian surface patch as:  
\begin{equation}
\mathbf{G}_p^\mathrm{surf} = \{\mathbf{x}_p, \mathcal{A}_p, \mathbf{N}_p, \boldsymbol{\sigma}_p, \mathbf{c}_p\}
\end{equation}
where $\mathbf{x}_p$ is the position, $\boldsymbol{\sigma}_p$ the opacity, and $\mathbf{c}_p$ the rendering color (see \cref{Sec:shading}). The new surface-specific attributes are the patch area $\mathcal{A}_p$ and surface normal $\mathbf{N}_p$.  

The surface area is derived from the Gaussian scaling $\mathbf{S}_p = \{\mathbf{S}_p^1, \mathbf{S}_p^2, \mathbf{S}_p^3\}$, ordered from largest to smallest:  
\begin{equation}
\label{eq:gaussian_surface_area}
\mathcal{A}_p = \pi \frac{\mathbf{S}_p^1 \cdot \mathbf{S}_p^2 \cdot \mathbf{S}_p^3}{\min(\mathbf{S}_p^1, \mathbf{S}_p^2, \mathbf{S}_p^3)} = \pi \mathbf{S}_p^1 \mathbf{S}_p^2
\end{equation}

The initial surface normal $\mathbf{N}_p$ is defined along the smallest axis~\cite{guedon2023sugar}.

\begin{figure}[!bt]
\centering
\includegraphics[width=\linewidth]{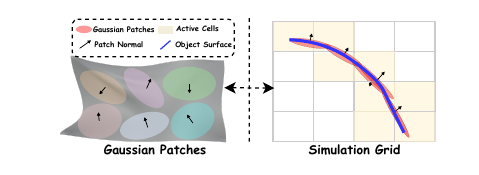}

\caption{\textbf{Surface Formulation.} 
\OURS{} represents objects as Gaussian surface patches, each defined by a normal and effective area, enabling efficient surface-specific interactions. 
Left: a deforming flag modeled as Gaussian patches. 
Right: coupling with the MPM grid, where patches provide surface attributes and receive deformation updates. 
Blue line indicates the ground-truth surface, and shaded cells mark active simulation regions.}
\label{fig:comparison}
\end{figure}

During simulation, a Gaussian patch evolves in world space under local deformation. Its center and covariance update according to the deformation map $\phi(\mathbf{X}_p,t)$ and gradient $\mathbf{F}_p(t)$:
\begin{equation}
\label{update_x_a}
    \mathbf{x}_p(t) = \phi(\mathbf{X}_p,t), \quad
    \mathbf{a}_p(t) = \mathbf{F}_p(t)\mathbf{A}_p\mathbf{F}_p(t)^T
\end{equation}
assuming local affine transformations as in~\cite{xie2024physgaussian}. The surface normal is updated as:
\begin{equation}
\label{update_normal}
    \mathbf{n}_p(t) = \frac{\mathbf{F}_p(t)^{-T}\mathbf{N}_p}{\Vert \mathbf{F}_p(t)^{-T}\mathbf{N}_p \Vert}
\end{equation}

As illustrated in \cref{fig:comparison}, Gaussian patches approximate object surfaces without requiring meshing. Each patch provides area and normal information while being coupled with the MPM grid, forming an efficient bridge between surface geometry and physical dynamics. This surface-based formulation enables accurate aerodynamic force modeling and efficient shading within a unified framework.

\subsubsection{Training}
During training, we encourage Gaussians to stay close to object surfaces and maintain surface fidelity under deformation. To this end, we adopt the anisotropy loss~\cite{xie2024physgaussian,feng2024gaussian} and entropy loss~\cite{duan20244d}:
\begin{equation}
    \mathcal{L}_{\mathrm{aniso}} = \frac{1}{n} \sum_{i=1}^n \max \left\{ \frac{S_i^1}{S_i^2} - a, 0 \right\}
\end{equation}

\begin{equation}
    \mathcal{L}_{\mathrm{entropy}} = -\frac{1}{n} \sum_{i=1}^n \sigma_i \log \sigma_i
\end{equation}
with $a = 1.1$. The anisotropy loss prevents Gaussians from becoming overly elongated, while the entropy loss drives opacities toward $0$ or $1$, pushing Gaussians to align with surfaces.  

We further introduce a size loss to constrain Gaussian scale within surface ranges:
\begin{equation}
    \mathcal{L}_{\mathrm{size}} = \frac{1}{n} \sum_{i=1}^n \max \left\{ S_i^1 - b, 0 \right\}
\end{equation}
where $b = 0.008$. The final training objective is:  
\begin{equation}
    \mathcal{L}_{\mathrm{tot}} = \mathcal{L}_{\mathrm{rgb}} + \lambda_1 \mathcal{L}_{\mathrm{aniso}} + \lambda_2 \mathcal{L}_{\mathrm{entropy}} + \lambda_3 \mathcal{L}_{\mathrm{size}}
\end{equation}
where $\mathcal{L}_{\mathrm{rgb}}$ is the standard RGB reconstruction loss from~\cite{kerbl3Dgaussians}. We set $\lambda_1 = 10$, $\lambda_2 = \lambda_3 = 0.01$. Further implementation details are provided in the supplementary material.

\begin{figure}[!htb]
\centering
\includegraphics[width=\linewidth,trim=0.5em 0 0.5em 0,clip]{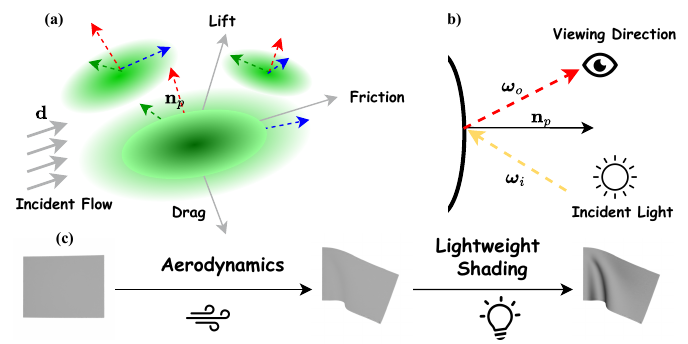}
\caption{\textbf{Illustration of Aerodynamics and Lightweight Shading.} 
\textbf{(a)} Incident flow (gray) and resulting aerodynamic forces on a Gaussian patch, with coordinate axes shown in red, green, and blue. 
\textbf{(b)} Principle of lightweight shading: the full BSDF formulation is approximated by an efficient local shading model. 
\textbf{(c)} Flag simulation results: (1) aerodynamics only, and (2) aerodynamics with lightweight shading.}
\label{fig:s_mpm}
\end{figure}

\subsubsection{Aerodynamics}

Aerodynamic forces arise from surface pressure and shear stress distributions governed by the Navier-Stokes equations~\cite{anderson2011ebook}:  
\begin{equation}
\mathbf{F} = \int_{A} \big( -p \mathbf{n} + \boldsymbol{\tau} \cdot \mathbf{n} \big) \, dA
\end{equation}
where $p$ is pressure, $\boldsymbol{\tau}$ is viscous stress, and $\mathbf{n}$ is the surface normal.

In practice, this formulation is commonly approximated by the classical dynamic pressure model through non-dimensionalization:
\begin{equation}
\mathbf{F} = \frac{1}{2} \rho V^2 A \, C(\mathbf{n}, \mathbf{d})
\end{equation}
where $\rho$ is fluid density, $V$ the relative velocity, $A$ the effective surface area, and $C$ a non-dimensional coefficient depending on the surface normal $\mathbf{n}$ and flow direction $\mathbf{d}$.

For a Gaussian surface patch $p$, the aerodynamic force is modeled as the combination of drag, friction, and lift:
\begin{equation}
\mathbf{f}_p
= \frac{1}{2} \rho v_{r,p}^2 A_p
\big( C_D \mathbf{n}_p + C_F \hat{\mathbf{d}}^{t}
+ C_L (\mathbf{d} \times \mathbf{n}_p) \big)
\end{equation}
where subscript $p$ indicates patch-specific quantities. $\hat{\mathbf{d}}^{t}$ is the tangential projection of the flow direction $\mathbf{d}$ onto the local surface tangent plane. The aerodynamic coefficients $C_D,C_F,C_L$ represent the drag, friction, and lift responses of each surface patch under incident flow. We follow standard practice in computational fluid dynamics. In implementation, we fix them for to representative values obtained from established empirical tables (please refer to supplementary materials). An illustration is provided in \cref{fig:s_mpm}~(a).

The characteristic swaying effect arises from fluctuations in flow strength and direction. To capture this behavior and improve visual realism, we perturb the incident velocity during simulation as
\begin{equation}
v' = v (1 + \epsilon), \quad \epsilon \sim \mathcal{U}(-\delta, \delta)
\end{equation}
where $\delta$ controls the perturbation strength and $\mathcal{U}(-\delta, \delta)$ is a uniform distribution.

\subsubsection{Lightweight Shading}
\label{Sec:shading}

The outgoing radiance at a surface point $p$ in direction $\omega_o$ is defined by the rendering equation with the bidirectional scattering distribution function (BSDF):
\begin{equation}
L_o(p, \omega_o) = \int_{\Omega} f_{\mathrm{BSDF}}(p, \omega_i, \omega_o)\, L_i(p, \omega_i)\, (\mathbf{n}_p \cdot \omega_i)\, d\omega_i
\end{equation}
where $f_{\mathrm{BSDF}}$ describes material response and $L_i$ is incident radiance from direction $\omega_i$.

To achieve real-time efficiency, graphics systems often approximate the BSDF with local illumination models such as the Phong model, consisting of ambient, diffuse, and specular components:
\begin{equation}
L_o(p, \omega_o) = k_a L_a + k_d (\mathbf{n}_p \cdot \omega_i) L_i + k_s (\mathbf{r} \cdot \omega_o)^{\alpha}
\end{equation}
where $k_a, k_d, k_s$ are ambient, diffuse, and specular coefficients, $\alpha$ is the shininess exponent, and $\mathbf{r}$ is the reflection vector:  
\begin{equation}
\mathbf{r} = 2(\mathbf{n}_p \cdot \omega_i)\mathbf{n}_p - \omega_i
\end{equation}

Details are visualized in \cref{fig:s_mpm} (b). For typical cloth-like objects such as flags, specular effects are minimal. In these cases, we approximate shading with a diffuse formulation directly tied to Gaussian surface properties. Specifically, given a unit light vector $\mathbf{L}$, the diffuse shading term at timestamp $t$ is then:
\begin{equation}
\label{eqn:gaussian_color}
s_p(t) = \max(0, \mathbf{n}_p(t) \cdot \mathbf{L})
\end{equation}

The rendered color for a Gaussian patch $p$ is computed as
\begin{equation}
\label{eqn:gaussian_color_2}
\mathbf{c}_p(t) =
\mathbf{c}_{p,0}(t) s_p(t)
\end{equation}
where albedo and light intensity are absorbed into the intrinsic color $\mathbf{c}_{p,0}(t)$, which is obtained from the spherical harmonic coefficients $\mathcal{C}_p$.

This formulation yields smooth shading transitions across surfaces without requiring mesh reconstruction or UV unwrapping. As illustrated in \cref{fig:s_mpm} (c), shading becomes a natural extension of surface-based Gaussian tracking, enabling visually compelling yet lightweight rendering.

%% file: sec/4_experiments.tex
\section{Experiments}

\begin{figure*}[!ht]
\centering
\includegraphics[width=\linewidth,trim=9em 0 3em 0,clip]{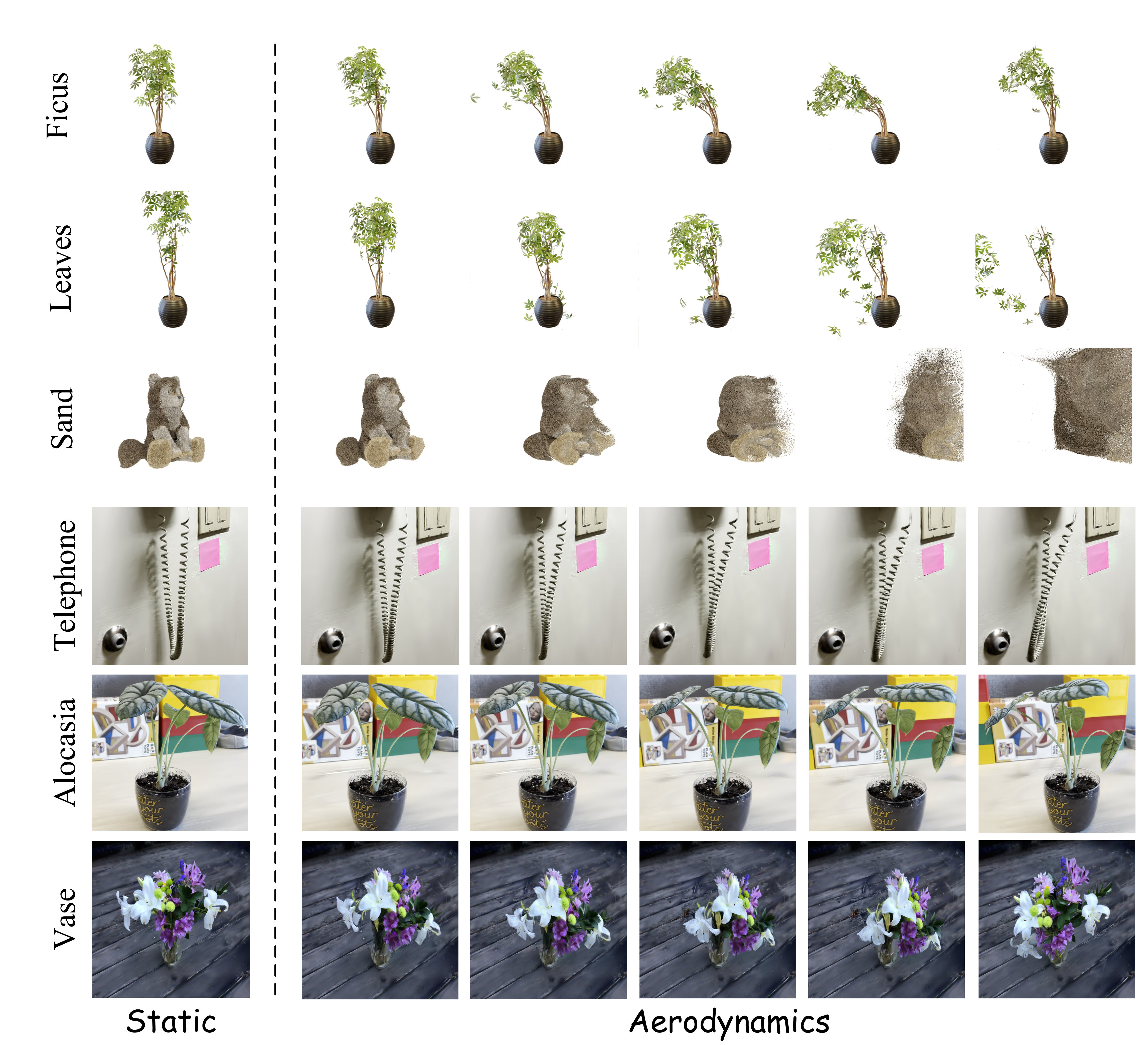}
\caption{\textbf{Qualitative Results of Aerodynamics.} 
\OURS{} generates realistic physics-based motion using 3D Gaussians across diverse materials.}
\label{fig:main}
\end{figure*}

We evaluate \OURS{} on synthetic and real-world datasets to assess its effectiveness, efficiency, and visual quality in simulating wind-driven dynamics. Comparative studies demonstrate advantages over state-of-the-art baselines, and ablation studies further validate the design choices.

\subsection{Evaluation of Aerodynamics}  

\paragraph{Datasets.}  
We test \OURS{} across diverse aerodynamic scenarios. Synthetic datasets include \textit{Ficus} and \textit{Sand}, generated by BlenderNeRF~\cite{Raafat_BlenderNeRF_2024}. Real-world scenes include \textit{Vase} from the NeRF dataset~\cite{mildenhall2020nerf}, and \textit{Telephone} and \textit{Alocasia} from PhysDreamer~\cite{zhang2024physdreamer}. For real-world scenes, initial point clouds and camera parameters are reconstructed using COLMAP~\cite{schonberger2016structure,schonberger2016pixelwise}.

\paragraph{Implementation Details.}
\OURS{} is implemented on top of the MPM engine from~\cite{xie2024physgaussian,zong2023neural}. Simulation regions, object materials, and parameters are manually specified, and Gaussian kernels with opacity below a threshold are filtered out. Wind speed, aerodynamic coefficients, and noise settings are detailed in the supplementary. All experiments, including Gaussian training and simulation, are performed on a single NVIDIA RTX 4090 GPU.  

\paragraph{Results.}  
Qualitative results are shown in \cref{fig:main}, with additional videos provided in the supplementary material. In \textit{Ficus} and \textit{Leaves}, \OURS{} captures realistic swaying and bending of branches, faithfully modeling surface-level wind interactions. In real-world cases such as \textit{Vase} and \textit{Alocasia}, our method produces visually convincing motion while preserving surface fidelity. Comparative qualitative results against baselines are included in the supplementary.

\begin{table*}[!ht]
	\centering
    \caption{\textbf{Quantitative Comparison on Flag Simulation.} 
    Metrics include PSNR, Chamfer Distance (CD), and Fréchet Video Distance (FVD), computed against reference videos. 
    Each pattern is evaluated over 250-frame sequences. Best results are highlighted in bold.}
    \label{tab:flag_results}
    \setlength{\tabcolsep}{8pt}
	\resizebox{\linewidth}{!}{
		\begin{tabular}{lccccccccc}
			\toprule
			& \multicolumn{3}{c}{Pattern-1}  & \multicolumn{3}{c}{Pattern-2}  & \multicolumn{3}{c}{Pattern-3} \\
			 Method & PSNR $\uparrow$ & CD $\downarrow$ & FVD $\downarrow$ & PSNR $\uparrow$ & CD $\downarrow$ & FVD $\downarrow$ & PSNR $\uparrow$ & CD $\downarrow$ & FVD $\downarrow$ \\
			\midrule

			NeRF-Editing~\cite{yuan2022nerf} 
			& 14.53 & 0.0540 & 413.39 & 15.94 & 0.0197 & 569.68 & 17.57 & 0.0228 & 382.69 \\
			Deforming-NeRF~\cite{xu2022deforming}
			& 15.53 & 0.0466 & 330.15 & 19.41 & 0.0245 & 430.01 & 17.33 & 0.0251 & 359.05 \\
		PAC-NeRF~\cite{li2023pacnerf}
			& 15.78 & 0.0526 & 275.85 & 20.95 & 0.0210 & 437.91 & 18.19 & 0.0212 & 437.71\\

            PIE-NeRF~\cite{feng2024pie}       
			& 17.73 & 0.0269 & 152.79 & 19.29 & 0.0111 & 383.25 & 14.13 & 0.0308 & 529.19 \\ 
            \midrule
            Spring-Gaus~\cite{zhong2025reconstruction} & 19.20 & 0.0288 & 140.01 & 21.31 & 0.0086 & 359.86 & 14.92 & 0.0196 & 556.86 \\
            PhysGaussian~\cite{xie2024physgaussian} 
            & 19.75 & 0.0215 & 158.16 & 21.96 & 0.0054 & 327.68 & 14.44 & 0.0263 & 628.58 \\
		\rowcolor[gray]{.9} \OURS{} ({\bf ours})  & \textbf{21.44} & \textbf{0.0178} & \textbf{50.47} & \textbf{22.26} & \textbf{0.0052} & \textbf{290.21} & \textbf{18.60} & \textbf{0.0066} & \textbf{219.94} \\

			\bottomrule
		\end{tabular}
	}
\end{table*}

\subsection{Flag Simulation}

\paragraph{Settings and Dataset.}  
To evaluate the ability of \OURS{} to handle complex and flexible surfaces, we curate a custom \textbf{Flag Simulation Dataset} with BlenderNeRF~\cite{Raafat_BlenderNeRF_2024}. We use 200 images for scene reconstruction and reserve one unseen view for evaluation, aligned collinearly with the initial normals to satisfy \cref{eqn:gaussian_color}. Three distinct motion patterns are generated, with simulation parameters (\eg, material density, flow intensity, \etc) matched between \OURS{} and BlenderNeRF. 

\noindent\textbf{Pattern-1}: Flag pinned on the left edge and driven only by gravity. The flag is pure white to highlight shading effects.

\noindent\textbf{Pattern-2}: Flag anchored at three corners, with the bottom-left corner free. Motion is driven by gravity and incident flow to the right.

\noindent\textbf{Pattern-3}: Large flag suspended from the top, driven by gravity and rightward flow.  

Lightweight shading is enabled for all patterns, under a single area light oriented perpendicular to the flag plane. For evaluation, all baselines use identical material parameters and flow strength. The background is set to black to remove external influences.

\begin{figure}[!ht]
\centering
\includegraphics[width=\linewidth,trim=0.5em 0 1em 0,clip]{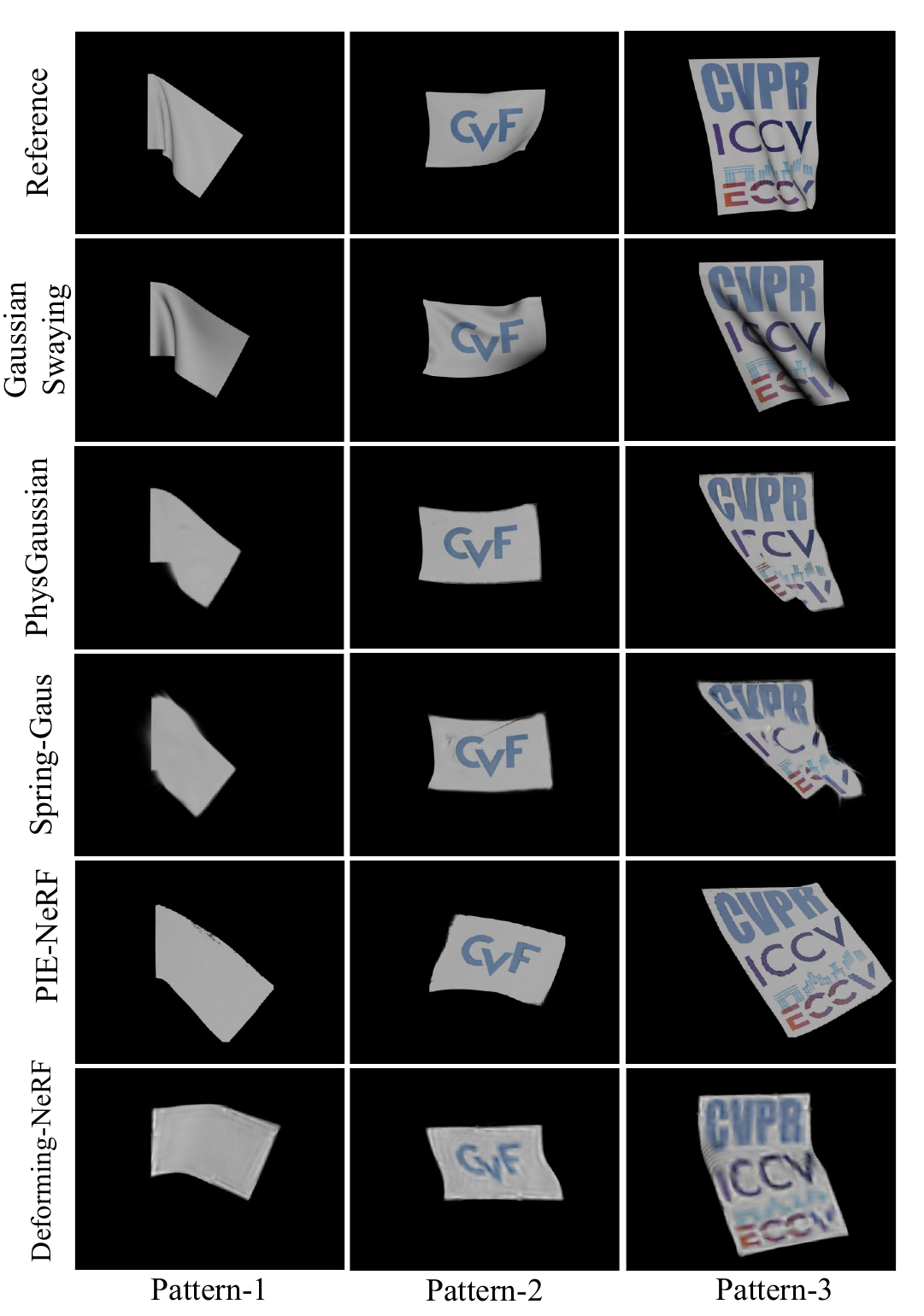}
\caption{
\textbf{Qualitative Comparisons on Flag Simulation Dataset}. The results are displayed on testing views.
}
\label{fig:main_flag}
\end{figure}

\paragraph{Quantitative and Qualitative Comparisons.}  
We compare \OURS{} against state-of-the-art frameworks for dynamic deformation with neural scene representations. NeRF-based baselines include NeRF-Editing~\cite{yuan2022nerf}, Deforming-NeRF~\cite{xu2022deforming}, PAC-NeRF~\cite{li2023pacnerf}, and PIE-NeRF~\cite{feng2024pie}. Gaussian-based baselines include PhysGaussian~\cite{xie2024physgaussian} and Spring-Gaus~\cite{zhong2025reconstruction}.

Evaluation metrics are Peak Signal-to-Noise Ratio (PSNR)~\cite{hore2010image}, Chamfer Distance (CD)~\cite{fan2017point}, and Fréchet Video Distance (FVD)~\cite{unterthiner2018towards}, computed using official implementations of the baselines. Each pattern is simulated for 250 frames at $1920 \times 1080$ resolution.

As shown in \cref{tab:flag_results}, \OURS{} achieves state-of-the-art performance across all metrics, with significant improvements in both reconstruction fidelity and temporal coherence. Qualitative results in \cref{fig:main_flag} further highlight the advantages: \OURS{} produces smooth shading transitions and realistic surface ripples consistent with reference videos. In contrast, particle-based methods (PIE-NeRF, PhysGaussian, PAC-NeRF, Spring-Gaus) struggle to capture surface-specific dynamics, while mesh-based methods (Deforming-NeRF, NeRF-Editing) require manual vertex manipulation and suffer from coarse reconstructions and surface discrepancies.

\subsection{Ablation Studies and Analysis}
\label{Sec:Ablation}

We conduct ablations to validate the design of \OURS{} and analyze its efficiency.

\begin{figure}[!ht]
\centering
\includegraphics[width=\linewidth]{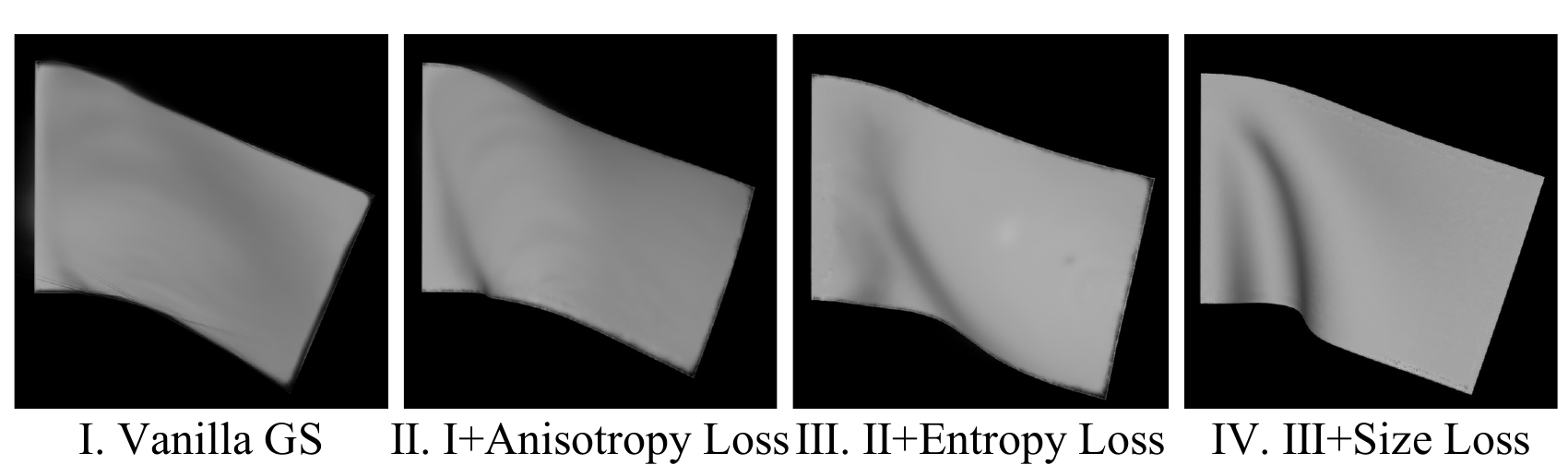}
\caption{\textbf{Ablation on Training Losses for 3D Gaussians.} 
Anisotropy and entropy losses improve rendering quality over vanilla Gaussian Splatting but still yield blurred regions. 
Adding size loss resolves this issue, producing realistic surface ripples.}
\label{fig:ablation_qualitative}
\end{figure}

\noindent\textbf{Training Losses.}  
We first study the impact of the proposed training losses. As shown in \cref{fig:ablation_qualitative}, the baseline (\uppercase\expandafter{\romannumeral1}) produces blurred surfaces. Adding anisotropy and entropy losses (\uppercase\expandafter{\romannumeral2}, \uppercase\expandafter{\romannumeral3}) improves sharpness but leaves artifacts. Incorporating size loss (\uppercase\expandafter{\romannumeral4}) yields clear surface ripples and the most realistic results. We also provide ablation studies on choices of hyperparameters in supplementary materials.

\begin{figure}[!ht]
\centering
\includegraphics[width=\linewidth]{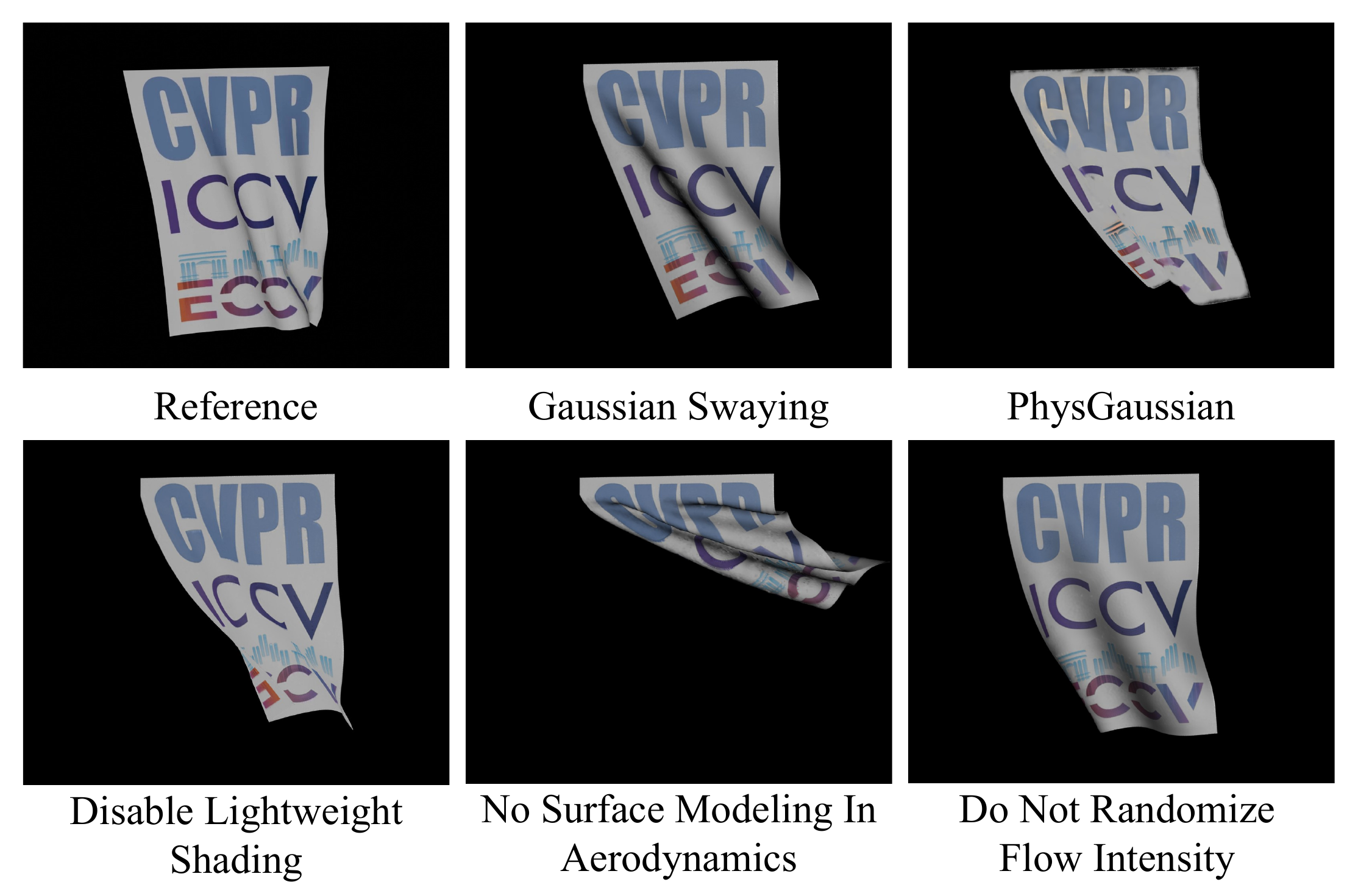}
\caption{\textbf{Qualitative Comparisons and Ablation Studies.} 
\textit{Top row:} comparison of \OURS{} with PhysGaussian and reference video.
\textit{Bottom row:} ablation of key design choices in \OURS{}, illustrated using Pattern-3.}
\label{fig:flag_component_ablations}
\label{fig:nvs_qualitative}
\end{figure}

\noindent\textbf{Design Components.}  
We ablate the key components of \OURS{}, with results summarized in \cref{tab:results_ablation_component} and visualized in \cref{fig:flag_component_ablations}.  
\begin{itemize}
    \item \textbf{No Flow Randomization.} Disabling randomness in flow intensity leads to unnatural motion (FVD $+114\%$).  
    \item \textbf{No Lightweight Shading.} Removing shading has little impact on dynamics ($+0.11\%$), but eliminates surface ripples (see \cref{fig:flag_component_ablations}), reducing visual realism.  
    \item \textbf{No Surface Modeling.} Applying aerodynamic forces directly on Gaussians without surface formulation causes severe degradation (FVD $+335\%$, CD $+329\%$, PSNR $-20.4\%$), performing even worse than PhysGaussian.  
\end{itemize}

\begin{table}[!ht]
	\centering
    \caption{\textbf{Quantitative Ablation on Design Choices in \OURS{}.} 
    Comparison with PhysGaussian using PSNR, Chamfer Distance (CD), and Fr\'{e}chet Video Distance (FVD).}
    \label{tab:results_ablation_component}
	\resizebox{\linewidth}{!}{
		\begin{tabular}{ccccc}
        \toprule
        Description & PSNR $\uparrow$ & CD $\downarrow$ & FVD $\downarrow$ \\ \midrule
            PhysGaussian  & 14.44 & 0.0263 & 646.51 \\
         \OURS{} ({\bf ours})& \bf 18.60 & \bf 0.0066 & \bf 219.94 \\
             \midrule
            Do Not Randomize Flow Intensity & 17.93 & 0.0070 & 468.72 \\
            Disable Lightweight Shading & 18.38 & 0.0069 & 243.57 \\
            W/o Surface Modeling in Aerodynamics & 14.80 & 0.0277 & 954.36 \\
         \bottomrule
		\end{tabular}
	}
\end{table}

\begin{figure}[!ht]
\centering
\includegraphics[width=0.9\linewidth]{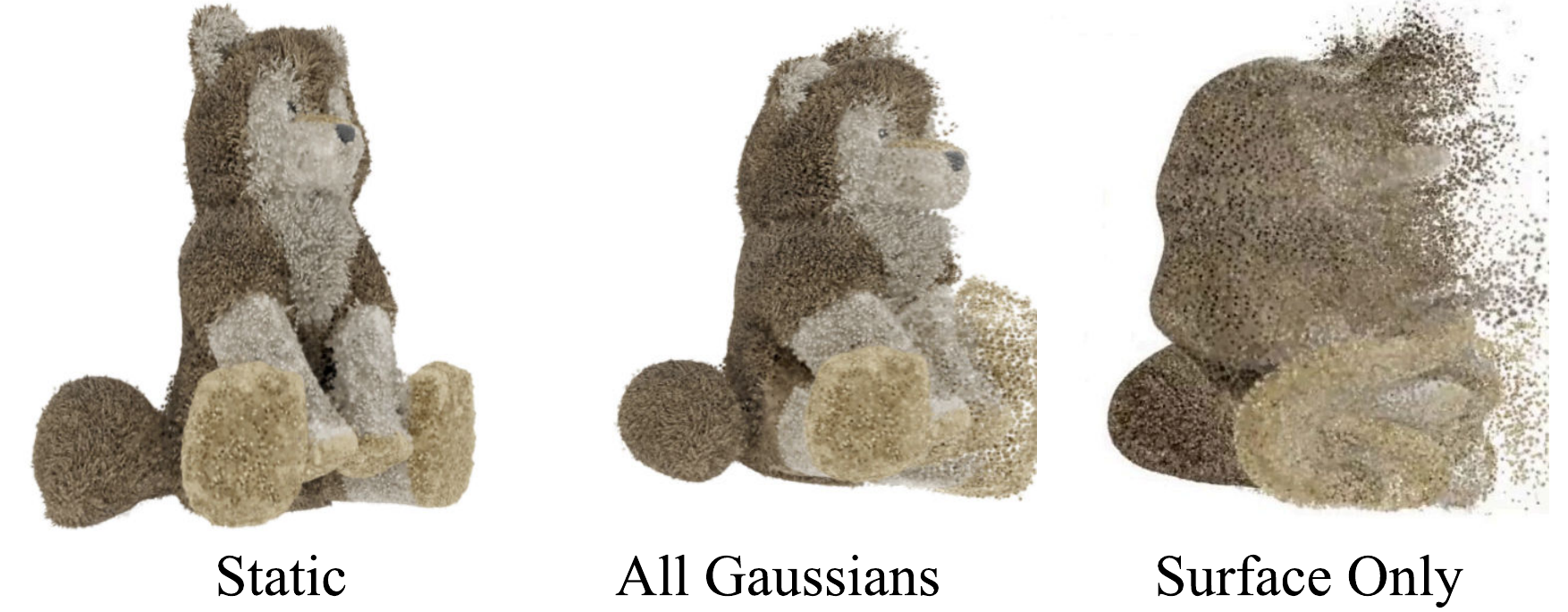}
\caption{\textbf{Simulations with different recipients of aerodynamic forces.}
}
\label{figure:surface}
\end{figure}

\noindent\textbf{Solid Objects.}  
We also study solid objects by filling their interior with Gaussians of identical density and volume (\cref{figure:surface}). Since \OURS{} applies aerodynamic forces only to surface Gaussians, internal Gaussians remain unaffected. This confirms that surface-aware modeling is essential for aerodynamics.

\noindent\textbf{Model Efficiency.}  
We compare per-frame runtime (simulation + rendering) and GPU memory on the flag simulation dataset. As shown in \cref{tab:efficiency}, \OURS{} achieves significant improvements, reducing runtime by $47.8\%$ and memory by $35.1\%$ compared to PIE-NeRF. Compared with PhysGaussian, \OURS{} is $25\%$ faster and uses $29.4\%$ less memory. 

\begin{table}[!ht]
	\centering
	\caption{\textbf{Efficiency Comparisons.} 
    Per-frame processing time and GPU memory usage on the flag simulation dataset.}
    \label{tab:efficiency}
	\resizebox{\linewidth}{!}{
		\begin{tabular}{cccc}
			\toprule
			Method & Time (s) $\downarrow$ & GPU Memory (GB) $\downarrow$ \\ \midrule
                Deforming-NeRF & 20.01 & 36.3 \\
			    PIE-NeRF & 0.23 & 7.4  \\
                PhysGaussian & 0.16 & 6.8 \\
                \rowcolor[gray]{.9} \OURS{} ({\bf Ours})  & \textbf{0.12} & \textbf{4.8} \\
			 \bottomrule
		\end{tabular}
	}
\end{table}

\noindent\textbf{Discussion.} Our proposed surface-aware modeling offers a continuous representation that avoids volumetric redundancy and gaps, reducing both storage for Gaussians and per-step simulation cost. Moreover, the unified pipeline enables the same Gaussian patches to support aerodynamic force computation and rendering, eliminating the need for separate primitives. These design choices directly contribute to the efficiency gains observed in our experiments, underscoring the practicality of \OURS{} for large-scale and real-time applications.

%% file: sec/5_conclusion.tex
\section{Conclusion}

We presented \OURS{}, a surface-based framework for aerodynamic simulation with 3D Gaussians. By modeling Gaussian patches as surface representations, \OURS{} unifies aerodynamic simulation and rendering without requiring meshing. Aerodynamic forces are computed per patch, while lightweight shading provides efficient, realistic appearance. Experiments on synthetic and real-world datasets, including a custom flag simulation benchmark, demonstrate that \OURS{} achieves state-of-the-art efficiency and visual fidelity.

%% file: sec/acknowledgment.tex
\section*{Acknowledgment} This work is supported by NSF award IIS-2127544 and NSF award IIS-2433768.

%% file: sec/supplementary.tex
\begin{table*}[!ht]
    \setlength{\tabcolsep}{1em}
	\centering
    \caption{\textbf{Parameter Settings.} We list the parameter settings on scenes in \OURS{}.}
    \label{tab:parameters}
	\resizebox{\linewidth}{!}{
		\begin{tabular}{lcccccccc}
        \toprule
		\multirow{2}{*}{\vspace{-3pt}Scene}	& \multicolumn{4}{c}{Material} & \multicolumn{4}{c}{Aerodynamics} \\
        \cmidrule(lr){2-5} \cmidrule(lr){6-9}
          & Constitutive Model & \textit{E} & $\nu$ & $\rho$ & $\mathbf{C}_{D}$ & $\mathbf{C}_{F}$ & $\mathbf{C}_{L}$ & Flow Intensity \\
        \midrule
        \textit{Ficus} (Truck) & Fixed Corotated & 2e6 & 0.4 & 300 & 0.5 & 0.4 & 0.01 & $(1.7,0,0)$\\
        \textit{Leaves} & Fixed Corotated & 1e4 & 0.4 & 15 & 0.5 & 0.4 & 0.01 & $(1.7,0,0)$ \\
        \textit{Sand} & Drucker-Prager & 5e5 & 0.3 & 200 & 0.4 & 0.3 & 0.01 & $(10,0,0)$\\
        \textit{Telephone} & Fixed Corotated & 5e5 & 0.4 & 200 & 0.5 & 0.4 & 0.01 & $(-1,0,1)$\\
        \textit{Alocasia}  & Fixed Corotated & 2e6 & 0.4 & 300 & 0.5 & 0.4 & 0.01 & $(0,-0.2,0)$\\
        \textit{Vase} & Fixed Corotated & 1e4 & 0.3 & 20 & 0.4 & 0.3 & 0.005 & $(0,0,-1.5),(0,0,1.5)$\\
        \textit{Flag} (Pattern-1) & Fixed Corotated & 5e3 & 0.3 & 50 & - & - & - & $(0,0,0)$ \\
        \textit{Flag} (Pattern-2) & Fixed Corotated & 3e3 & 0.3 & 30 & 0.1 & 0.3 & 0.005 & $(2.5,0.5,0)$ \\
        \textit{Flag} (Pattern-3) & Fixed Corotated & 3e3 & 0.3 & 20 & 0.1 & 0.3 & 0.005 & $(2,0,0)$ \\
        \bottomrule
		\end{tabular}
	}
\end{table*}

\begin{figure*}[!bt]
\centering
\includegraphics[width=\linewidth]{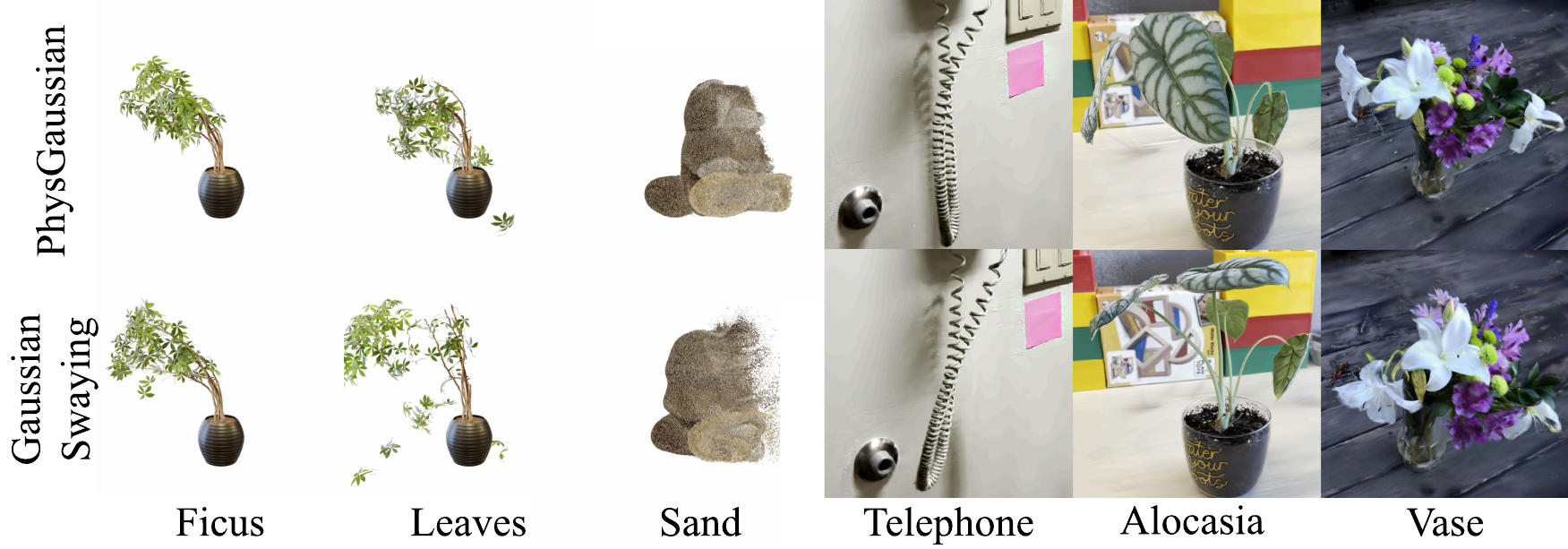}
\caption{Qualitative comparisons between PhysGaussian (\textit{Upper Row}) and Gaussian Swaying (\textit{Lower Row}).
}
\label{fig:quali_comparisons}
\end{figure*}

\section{MPM Algorithm}
MPM simulates dynamic movements via discretizing a continuous body into particles and then updating their properties (including positions, velocities, etc.). We summarize the MPM algorithm as follows:  

\begin{enumerate}
    \item \textbf{Transfer Particles to Grid.} The mass and momentum from particles are transformed to grid nodes as 
    \begin{equation}
    \begin{aligned}
        m_{{i}}^n &=\sum_p w_{{i} p}^n m_p, \\
        m_i^n\mathbf{v}_{{i}}^n &= \sum_p w_{{i} p}^n m_p\left(\mathbf{v}_p^n+\boldsymbol{C}_p^n\left(\mathbf{x}_{{i}}-\mathbf{x}_p^n\right)\right),
    \end{aligned}
    \end{equation}
    where $\boldsymbol{C}_p^n$ refers to affine momentum~\cite{jiang2015affine} on particle $p$.
    \item \textbf{Grid Update.} Update grid velocities based on forces at the next timestep by 
    \begin{equation}
        \frac{m_i}{\Delta t}(\mathbf{v}_i^{n+1} - \mathbf{v}_i^n) = - \sum\limits_p \tau_p^n  \nabla w_{ip}^n V_p^0 + \mathbf{f}_i^{ext},
    \end{equation}
    \item \textbf{Transfer Grid to Particles.} Transfer velocities back to particles and update particle states. 
    \begin{equation}
    \begin{aligned}
        \mathbf{v}_p^{n+1}&=\sum_i \mathbf{v}_{{i}}^{n+1} w_{{i} p}^n, \\
        \mathbf{x}_p^{n+1} &= \mathbf{x}_p^{n} + \Delta t \mathbf{v}_p^{n+1}, \\
        {\boldsymbol{C}}_p^{n+1}&=\frac{12}{\Delta x^2(b+1)} \sum_{{i}} w_{{i} p}^n \mathbf{v}_i^{n+1} \left(\mathbf{x}_{{i}}^n-\mathbf{x}_p^n\right)^T,\\
        \nabla \mathbf{v}_p^{n+1} &= \sum_i \mathbf{v}_{{i}}^{n+1} {\nabla w_{{i} p}^n}^{T},\\
        \mathbf{F}_p^{E,n+1} &= \mathcal{Z}((\mathbf{I} + \nabla \mathbf{v}_p^{n+1}) \mathbf{F}^{E, n}), \\
        tau_p^{n+1} &= \boldsymbol{\tau}(\mathbf{F}_p^{E, n+1}).
    \end{aligned}
    \end{equation}
    Here $b$ is the B-spline degree, and $\Delta x$ is the Eulerian grid spacing. The deformation gradient $\mathbf{F}_p = \mathbf{F}_p^E \mathbf{F}_p^P$ is decomposed into elastic part $\mathbf{F}_p^E$ and plastic part $\mathbf{F}_p^P$.
    Detailed computation of the return map $\mathcal{Z}$ and the Kirchhoff stress $\boldsymbol{\tau}$ is listed in \cref{Sec:elastic model}. Please refer to \cite{jiang2016material,zong2023neural} for more details.
\end{enumerate}

\section{Details on Elastic and Plastic Models}
\label{Sec:elastic model}
In this section, we elaborate on the elastic and plastic constitutive models used in \OURS{}, as listed in \cref{tab:parameters}. We adopt the models from \cite{zong2023neural}.

\subsection{Fixed Corotated Elasticity}
The Kirchhoff stress $\tau$ is defined as 
\begin{equation}
\tau_p = 2  \mu(\mathbf{F}^E_p-\mathbf{R}_p) {\mathbf{F}_p^E}^{T}+\lambda(J_p-1) J_p,
\end{equation}
where $\mathbf{R}_p = \mathbf{U}_p \mathbf{V}_p^T$ denotes local rotation and $\mathbf{F}_p^E = \mathbf{U}_p\mathbf{\Sigma_p}\mathbf{V}_p^T$ is the singular value decomposition of elastic deformation gradient. $J_p$ is the determinant of $\mathbf{F}^E_p$ \cite{jiang2015affine}.

\subsection{Drucker-Prager Plasticity}
Drucker-Prager plasticity is used for simulation of sand \cite{klar2016drucker}.  

We set friction angle as $\phi_f$ and $\alpha=\sqrt{\frac{2}{3}} \frac{2 \sin \phi_f}{3-\sin \phi_f}$. We set $\phi_f=30^{\circ}$ in the experiment.
Then we calculate intermediary variants:
\begin{equation}
    \hat{\epsilon_p} = \epsilon_p - \frac{\text{tr}(\epsilon_p)}{d} \mathbf{I}, \; \delta \gamma=\|\hat{\epsilon_p}\|_F+\alpha \frac{(d \lambda+2  \mu) \text{tr}(\epsilon_p)}{2  \mu},
\end{equation}
where $\mathbf{F}_p^E = \mathbf{U}_p \mathbf{\Sigma}_p \mathbf{V}_p^T$ and $\boldsymbol{\epsilon}_p=\log (\boldsymbol{\Sigma}_p)$. $d$ is the dimension and $\hat{\epsilon_p}$ is the plastic deformation amount. 

The detailed calculation of return mapping $\mathcal{Z}$ is given as

\begin{equation}
    \mathbf{F}^E_p = \mathbf{U}_p  \mathcal{Z}(\boldsymbol{\Sigma}_p ) \mathbf{V}^T_p,
\end{equation}

Here, if $\text{tr}(\epsilon_p)>0$, then $\mathcal{Z}(\boldsymbol{\Sigma}_p ) = 1$. $\mathcal{Z}(\boldsymbol{\Sigma}_p ) = \boldsymbol{\Sigma}_p $ if $\hat{\epsilon_p} \leq 0 \text{ and } \text{tr}(\epsilon_p)\leq0$. For other conditions, $\mathcal{Z}$ is calculated as $\exp \left(\boldsymbol{\epsilon}_p-\delta \gamma _p\frac{\hat{\epsilon_p}}{\|\hat{\epsilon_p}\|}\right)$.

\section{More Details on Parameters}

We further provide detailed parameter settings, including Young’s modulus $\textit{E}$, Poisson’s ratio $\nu$, mass density $\rho$, aerodynamic force coefficients $\mathbf{C}_{D}$, $\mathbf{C}_{F}$, $\mathbf{C}_{L}$, flow intensity and constitutive model of material in \cref{tab:parameters}. The constitutive model is defined the same with \cite{xie2024physgaussian}. If the flow intensity changes during the simulation, it will be denoted by two values in the table (\textit{Cloth} and \textit{Vase} in \cref{tab:parameters}).

By default, we add three-dimension Gaussian noise with mean value 0 and standard deviation 0.3 to the wind speed to best approximate the behavior of the natural wind. The biggest component of the wind speed (e.g. 1.7 in \textit{Leaves} in \cref{tab:parameters}) also serves the mean value of a sine function. The amplitude of the sine function is set to half of this component's value to simulate the swaying effect. In the flag simulation dataset, gravity is set as $9.8m/s^2$.

We train 3D Gaussians using the official implementation of \cite{kerbl3Dgaussians} with proposed training losses. In Tab.~\ref{tab:fvd}, we showcase ablation studies on Gaussian size, where $b=0.008$ is optimal for our setting. The total iterations are 60,000. In the simulation, Gaussians are filtered by an opacity threshold of 0.1 (In \textit{Sand} we set it as 0.02). The frame number is 250, with each frame generated every 0.04s. The frame rate for the video is 25 for all experiments.

\begin{table}[!ht]
	\centering
	\caption{
	\bf Ablation studies on Gaussian size.
	}
        \label{tab:fvd}
	\resizebox{\linewidth}{!}{
		\begin{tabular}{ccccccc}
			\toprule
			Param $b$ & 0.001 & 0.005 & 0.008 & 0.01 & 0.05 & 0.1 \\ \midrule
                FVD $\downarrow$ & 362.77 & 248.66 & \textbf{219.94} & 432.88 & 796.01 & 910.24 \\
			 \bottomrule
		\end{tabular}
	}
\end{table}

\section{More Results}

We visualize qualitative comparisons with PhysGaussian in \cref{fig:quali_comparisons}. As shown, PhysGaussian fails to capture surface-level interaction (\eg, unrealistic leaf orientations in \textit{Ficus} and \textit{Leaves}, sand stretching rightward without scattering in \textit{Sand}).

\textit{Vase}, \textit{Telephone} and \textit{Alocasia} are real-world datasets to validate the application of \OURS{}.
In \textit{Telephone} and \textit{Alocasia}, objects are applied with a gentle leftward incident flow. For comparisons, PhysGaussian unnaturally overstretches alocasia and telephone line, while ours accurately models leaf orientation and telephone line floating in the wind.

\section{Object Editing}

\begin{figure}[!bt]
\centering
\includegraphics[width=\linewidth]{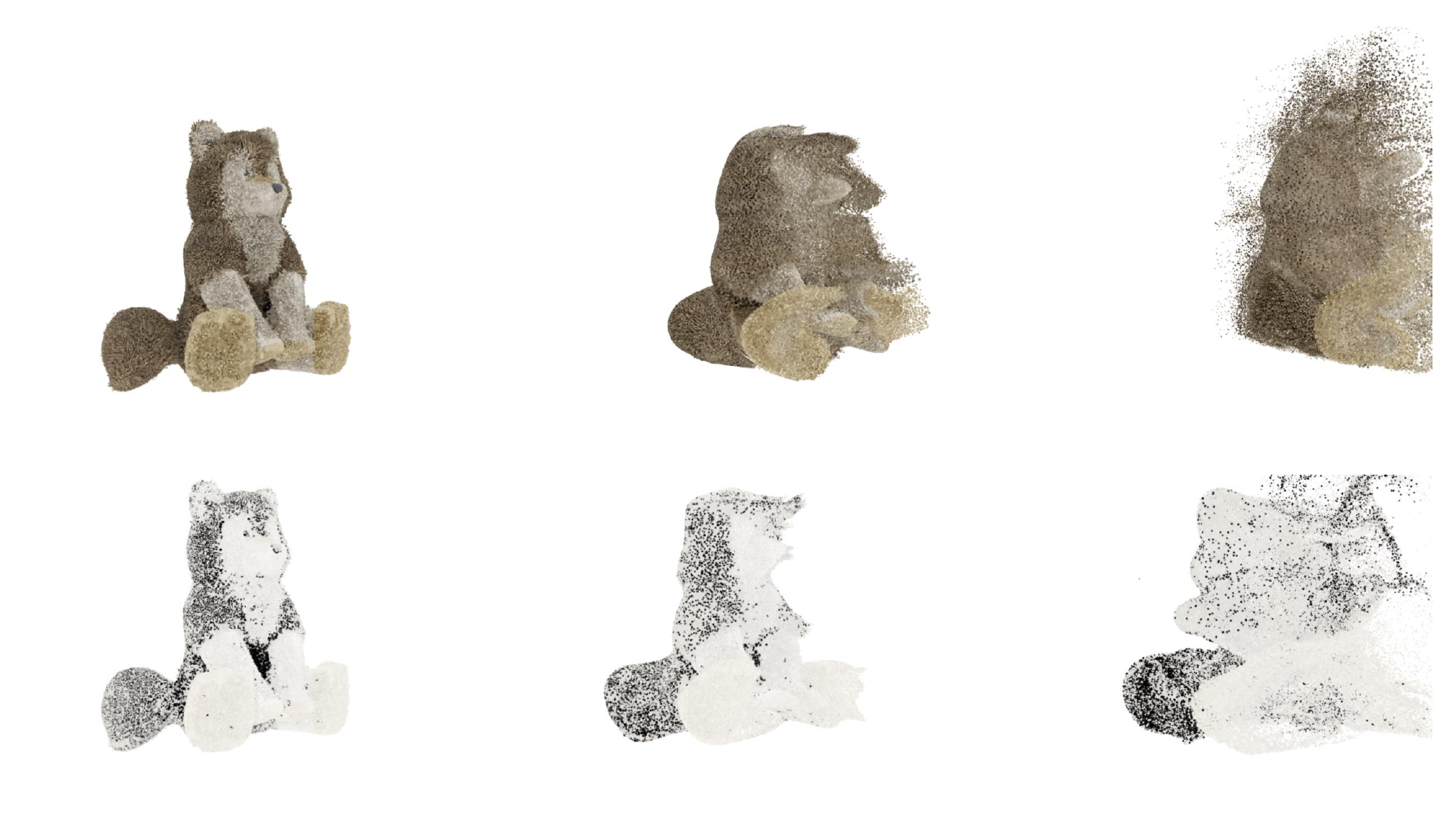}
\caption{
\textbf{Object Editing}. \OURS{} is capable of realistic material and color editing to facilitate versatile applications. The top sand bear is edited to white foam at the bottom.
}
\label{fig:editing}
\end{figure}

To further demonstrate the flexibility and versatility of Gaussian Swaying, we explore its capability for real-time scene editing, including adjustments to material properties, colors, and interaction parameters. In \cref{fig:editing}, the \textit{Upper Row} is \textit{Sand} while the \textit{Lower Row} is edited to \textit{Foam} with white color. Foam generally is stickier than sand, which aligns with the generated results by \OURS{}.

\section{Video Results}
We provide video results in the supplementary materials. For example, detailed surface-level interactions between objects and wind can be best reflected in `Ficus.mp4' and `Leaves.mp4'. In `Pattern3.mp4', surface ripples can be observed, which creates realistic visual effects. The reference videos of flag simulation are prefixed by ``Reference-''. Compared with `Sand.mp4', in `Sand\_All\_Gaussians.mp4' the solid object is moving directly without surface deformation, which demonstrated the necessity of surface modeling.